\def\year{2020}\relax
\documentclass[letterpaper]{article} 
\usepackage{aaai20}  
\usepackage{times}  
\usepackage{helvet} 
\usepackage{courier}  
\usepackage[hyphens]{url}  
\usepackage{graphicx} 
\usepackage{graphicx}  
\usepackage{bm}
\usepackage{subfigure}
\usepackage{booktabs}
\usepackage{threeparttable}
\usepackage{amssymb}
\usepackage{float}
\usepackage[colorlinks,linkcolor=blue]{hyperref}
\title{
	FFA-Net:	Feature Fusion Attention Network for Single Image Dehazing
}
\author{Xu Qin\textsuperscript{\rm 1*}
Zhilin Wang \textsuperscript{\rm 2}\thanks{The first two authors contributed equally to this work.}
Yuanchao Bai \textsuperscript{\rm 1}
Xiaodong Xie\textsuperscript{\rm 1}\thanks{Corresponding author}
Huizhu Jia \textsuperscript{\rm 1}\\ 
\textsuperscript{\rm 1}School of Electronics Engineering and Computer Science, Peking University\\
\textsuperscript{\rm 2}School of Computer Science and Engineering, Beihang University\\
\textrm{
\{qinxu,yuanchao.bai,donxie,hzjia\}@pku.edu.cn, 007wangzhilin@buaa.edu.cn}
}
 
\begin{document}

\maketitle

\begin{abstract}
In this paper, we propose an end-to-end feature fusion at-tention network (FFA-Net) to directly restore the haze-free image. 
The FFA-Net architecture consists of three  key components:

1) A novel Feature Attention (FA) module combines Channel Attention with Pixel Attention mechanism, considering that  different channel-wise features contain totally different weighted information  and haze distribution is uneven on the different image pixels.
FA treats different features and pixels unequally, which provides additional flexibility in dealing with different types of information, expanding the representational ability of CNNs.
2) A basic block structure consists of Local Residual Learning and Feature Attention, Local Residual Learning allowing the less important information such as thin haze region or low-frequency to be bypassed through multiple local residual connections, let main network architecture focus on more effective information. 
3) An Attention-based different levels Feature Fusion (FFA) structure, the feature weights  are adaptively learned from the Feature Attention (FA) module, giving more weight to important features. This structure can also retain the information of shallow layers and pass it into deep layers.

The experimental results demonstrate that our proposed  FFA-Net surpasses previous state-of-the-art single image dehazing methods by a 
very large margin both quantitatively and qualitatively, boosting the best 
published PSNR metric from 30.23 dB to  36.39 dB on the SOTS indoor test dataset.
 Code has been made available at
\href{https://github.com/zhilin007/FFA-Net}{GitHub}.

\end{abstract}

\section{Introduction}
Single image dehazing as a fundamental low-level vision task, has attracted increasing attention in the computer vision community and artificial intelligence companies over the past few decades.
\begin{figure}[t]
	\centering
	\includegraphics[width=0.95\columnwidth]{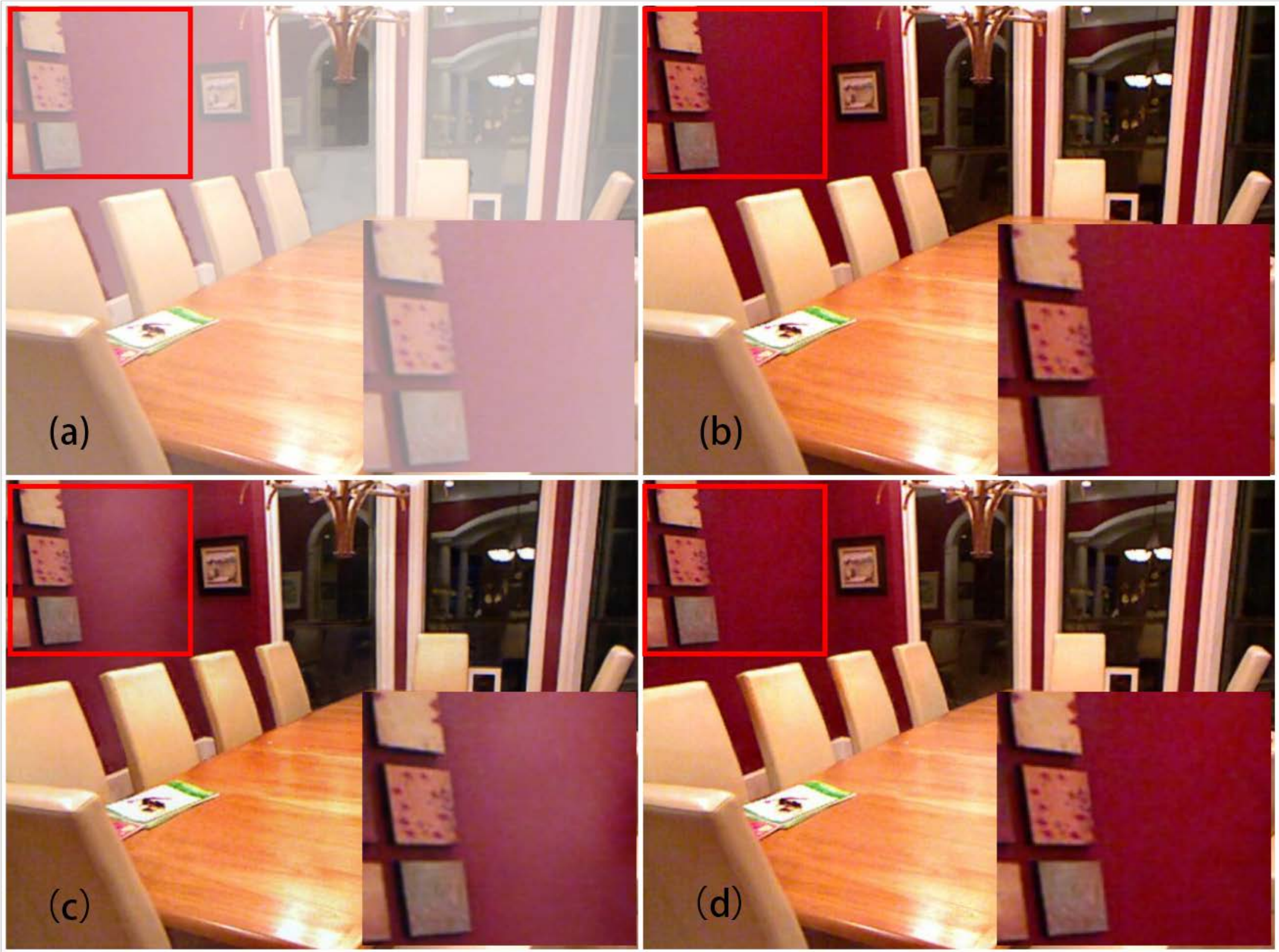} 
	\caption{An example of image dehazing. (a) Input hazy image.  (b) Ground Truth. (c) Result of GCANet. (d) Our result.  Our FFA-Net outperforms GCANet in image detail and color fidelity.}
	\label{fig1}
	
\end{figure}

Due to the existence of smoke, dust, fumes, mist and other floating particles in the atmosphere,
images taken in such atmosphere are often subject to  color distortion, blurring, low contrast and other visible quality degradation, and the hazy image input will make it difficult to solve the other visual tasks such as classification, tracking, person re-identification  and  object detection.
In view of this, image dehazing  aims to recover the clean image from the corrupted input, this will be the preprocessing step of the high-level vision tasks. The atmosphere scattering model \cite{scatteringmodel17}\cite{scatteringmodel20}\cite{scatteringmodel21} provides a simple approximation of the haze effect, it is formulated as: 
\begin{equation}
\bm{I}(z)=\bm{J}(z)t(z)+\bm{A}(1-t(z))
\label{form1}
\end{equation}

Where $\bm{I}(z)$ is the observed hazy image, $\bm{A}$ is the global atmosphere light, and $t(z)$ is the medium transmission map,
$\bm{J}(z)$ is the haze-free image. Moreover, we have $ t(z)=e^{-\beta d(z)} $with $\beta$ and $d(z)$ being the atmosphere scattering parameter and the scene depth, respectively. The atmosphere scattering model shows that image dehazing is an underdetermined problem without the knowledge of $\bm A$ and $t(z)$.
	Formulation (1) can also be formulated as:
$$ \bm J(z)=\frac{(I(z)-\bm A)}{t(z)}+A \eqno(2)$$
From the formulation 1 and 2, we can notice that if we estimate the global atmosphere and transmission map properly for a captured hazy image, we can restore a clear haze-free image.

Based on the atmosphere scattering model, early dehazing methods did a series of works\cite{featureforward6}\cite{featureforward10}\cite{featureforward11}\cite{featureforward15}\cite{featureforward17}\cite{featureforward23}\cite{featureforward34}.  DCP is one of the outstanding prior-based methods, they propose  the dark channel prior based on the assumption that image patches of outdoor haze-free images often have low-intensity values in at least one channel. However, the prior-based methods may lead to an inaccu-rate estimation of transmission map because of the prior may be easily violated in practice, so the prior-based meth-ods may not work well in certain real cases.

With the rising-up of deep learning, many neural network approaches have also been proposed to estimate the haze effect, including the pioneering work of DehazeNet\cite{dehazenet}, the multi-scale CNN(MSCNN)\cite{mscnn}, the residual learning technique\cite{resnet}, the quad-tree CNN\cite{quad_treeCNN-adaptive}, and the densely connected pyramid dehazing network\cite{dense_pyramid}. Compared to traditional  methods, deep learning methods try to directly regress the intermediate transmission map or the final haze-free image. With the big data being applied, they achieve  superior performance with robustness.

In this paper, we propose a novel end-to-end feature fusion network(denoted as “FFA-Net”) for single image dehazing.

Previous CNN-based image dehazing networks treat the channel-wise and pixel-wise feature equally, but haze is unevenly distributed across an image, the weight of the very thin haze should be significantly different from that of the thick haze region pixels.  Furthermore, DCP also finds that it is very common that some pixels have very low intensity in at least one color(RGB) channel, this further illustrates that different channel features have totally different weighted information. If we treat it equally, it will spend plenty of resources on unnecessary computations for less important information, the network will lack the ability to cover all the pixels and channels. Finally, it will greatly limit the representation of the network.

Since the attention mechanism\cite{att1}\cite{att2}\cite{att3}  has been widely used in the design of neural networks, it has played an important role in the performance of networks. Inspired by the work \cite{rcan}, we further design a novel feature attention (FA) module. The FA module combines the channel attention and pixel attention in channel-wise  and pixel-wise features,  respectively. FA treats different features and pixels unequally, which can provide additional flexibility in dealing with different types of information.

The emergence of ResNet\cite{resnet} has made it possible to train a very deep network. We adopt the idea of skip connection and the attention mechanism and design a basic block consisting of multiple local residual learning skip connections and feature attention. For one thing, the local residual learning allows the information of the thin haze region and low-frequency information to be bypassed through multiple local residual learning, making the main network learn more useful information. And channel attention further improves the capability of FFA-Net.

As the network goes deeper and deeper, shallow feature information is often difficult to preserve. In order to identify and fuse different level features, U-Net\cite{unet} and other networks strive to integrate the shallow and deep information. Similarly,  we propose an attention-based feature fusion structure (FFA), this structure can retain shallow information and pass it into deep layers. Most importantly, the FFA-Net gives different weights to different level features before feeding all features to feature fusion module, the weight is obtained by adaptive learning of the FA module.  It is much better than those that directly specified the weight.

To evaluate the performance of different image dehazing networks, peak-signal-to-noise-ratio (PSNR) and structure similarity index (SSIM) are commonly used to quantify dehazed image restoration quality. For human subjective assessment, we also provide plenty of network outputs from corrupted inputs. We validate the effectiveness of the FFA-Net on the widely used dehazing benchmark dataset RESIDE\cite{RESIDEbenchmarking}. Compared the PSNR and SSIM metrics with previous state-of-the-art methods. Experiments demonstrate that FFA-Net surpasses all the previous methods both qualitatively and quantitatively by a very large margin. Moreover, we conduct many ablation experiments to prove that our key components of FFA-Net have an excellent performance.

\begin{figure*}[t]
	
	\centering
	\includegraphics[width=0.9\textwidth]{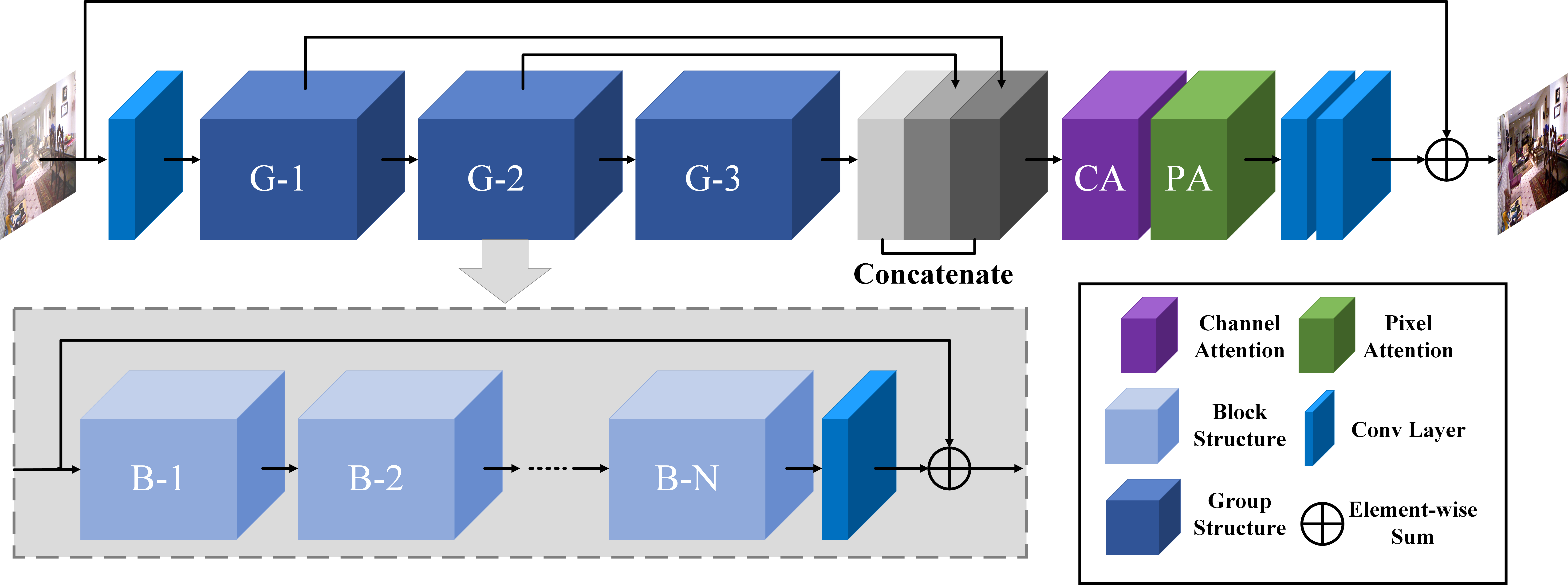} 
	\caption{The feature fusion attention network (FFA-Net) architecture.}
	\label{fig2}
	
\end{figure*}

Overall, our contributions are four-folds as below:
\begin{itemize}
\item We propose a novel end-to-end feature fusion attention network FFA-Net for single image dehazing. FFA-Net surpasses previous state-of-the-art image dehazing methods by a very large margin, the FFA-Net performs especially outstanding in region with thick haze and rich texture details. We also have a powerful advantage in  the restoration of image detail and color fidelity, as seen in Fig.\ref{fig1} and Fig.\ref{fig8}.

\item We propose a novel feature attention (FA) module, which combines the channel attention and pixel attention mechanism. This module
provides additional flexibility in dealing with different types of information, focusing more attention on the thick haze pixels and more important channel information. 

\item We propose a basic block consisting of local residual learning  and feature attention (FA), local residual learning allows the information of the thin haze region and low-frequency information to be bypassed through multiple skip connections, feature attention (FA) further improves the capacity of FFA-Net.

\item We propose an attention-based feature fusion (FFA) structure, this structure can retain shallow layers' information and pass it into deep layers.  Besides, it can not only fuse  all features but also adaptively learn the different weights of different level feature information. Finally, it achieves a much better performance than other feature fusion methods.

\end{itemize}

\section{Related Work}
Previously, most of the existing image dehazing methods depend on the formulation of physical scattering model equation\ref{form1}, which is a highly ill-posed problem because of the unknown transmission map and global atmospheric light. These methods can be roughly divided into two classes: traditional prior-based  methods and  modern learning-based methods. No matter which method is used, the key is to solve the transmission map and the atmosphere light.
For traditional methods, based on the different image statistics prior, they leverage it as extra constraints to compensate for the information loss during the corruption procedure. 

DCP\cite{featureforward11} proposed a dark channel prior for the estimation of the transmission map. However, the priors are found to be unreliable when the scene objects are similar to the atmospheric light. \cite{featureforward34} propose a simple but powerful color attenuation prior by creating a linear model for modeling the scene depth of the hazy image. \cite{gca11} present a new method for estimating the optical transmission in hazy scenes, the scattered light is eliminated to increase scene visibility and recover haze-free scene contrasts., \cite{featureforward6} proposed a non-local prior to characterize the clean image, the algorithm relies on the assumption that colors of a haze-free image are well approximated by a few hundred distinct colors, which forms tight clusters in RGB space.  Although these methods have made a series of success, the prior is not robust to handle all the cases, such as the unconstraint environment in the wild.

In view of the prevailing success of deep learning in image processing tasks and the availability of large image datasets,  \cite{dehazenet} proposed an end-to-end dehazing model based on convolution neural network DehazeNet, it takes
a hazy image as input, and outputs its medium transmission map, which is subsequently used to recover a haze-free image via atmospheric scattering model. \cite{mscnn}  employed a Multi-Scale MSCNN that is able to perform a  refined transmission map from the hazy image. \cite{proximal}  combines the advantages of traditional prior-based dehazing methods and deep learning methods by incorporating haze-related prior learning into deep network.. \cite{li2017aod}  AOD-Net directly
generates a clean image through a light-weight CNN. Such a novel end-to-end design makes it easy to embed AOD-Net into other deep models.
The gated fusion network(GFN) \cite{gfn}   leverages hand-selected  pre
processing methods and multi-scale estimation, which are generic in nature and are subject to improvement.
\cite{gca}  proposed an end-to-end gated context aggregation network to directly  restore the final haze-free image, which adopted the latest smoothed dilation technique to help remove the gridding artifacts caused by the widely used dilated convolution with negligible extra parameters. EPDN \cite{epdn} is embedded by a generative adversarial network, which is followed by a well-designed enhancer without relying on the physical scattering model.

\section{Fusion Feature Attention Network (FFA-Net) }
In this section, we mainly introduce our feature fusion attention network FFA-Net.  As shown in Fig.\ref{fig2},  the input of FFA-Net is a hazy image, it is passed  into a shallow feature extraction part,  then is fed  into N  Group Architectures  with multiple skip connections, the output features of N Group Architectures are fused together through our proposed Feature Attention module,  after that, the features  will be finally passed to the reconstruction part and global residual learning structure, thereby getting a haze-free output. 

Furthermore, every Group Architecture combines B Basic Blocks Architecture with local residual learning,  every Basic Block combines the skip connection and Feature Attention (FA) module.  FA is an  attention mechanism structure consisting of Channel-wise Attention and Pixel-wise Attention.

\subsection{Feature Attention (FA)}  		  
\begin{figure}[t]
	\centering
	\includegraphics[width=1\columnwidth]{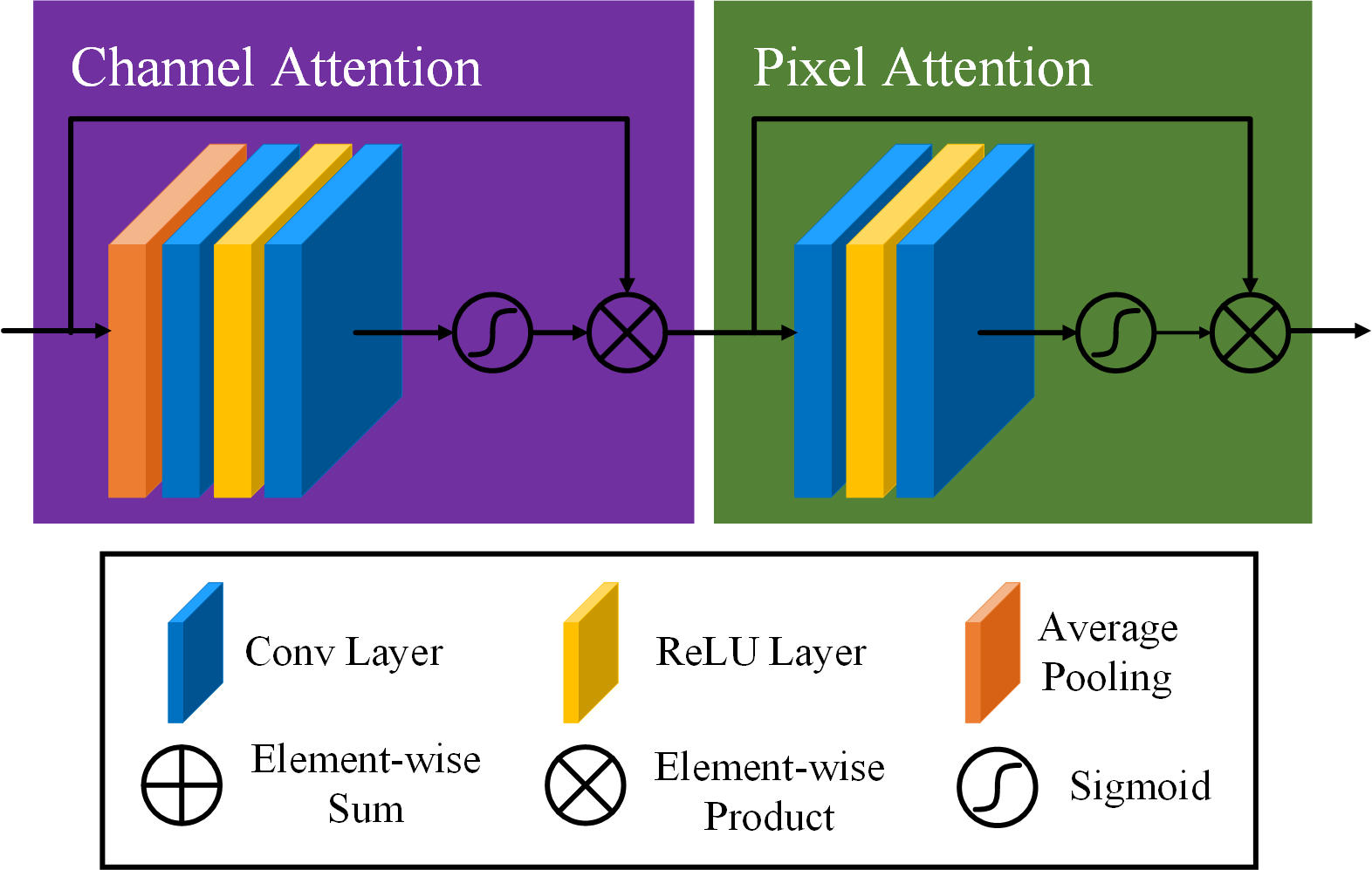} 
	\caption{ Feature Attention module}
	\label{fig3}
\end{figure}

Most image dehazing networks treat the channel-wise and pixel-wise features equally, which can not handle the image with uneven haze distribution  and weighted channel-wise feature properly. Our Feature Attention (see Fig.\ref{fig3}) consists of channel attention and pixel attention, which can provide additional flexibility in dealing with different types of information. 

FA treats different features and pixels region unequally, which can provide additional flexibility in dealing with different types 
of information, and can expand the representational ability of CNNs. 
The crucial step is how to generate different weights for each channel-wise and pixel-wise feature. Our solution is below.

\subsubsection{Channel Attention (CA)}

Our channel attention mainly concerns that different channel features have totally different weighted information with regards to DCP\cite{featureforward11}.
Firstly, we take the channel-wise global spatial information into a channel descriptor by using global average pooling. 

$$ g_{c}=H_{p}(F_{c})=\frac{1}{H\times W}\sum_{i=1}^{H}\sum_{j=1}^{W}X_{c}(i,j) \eqno(3)$$

Where $X_c(i,j)$  stands for the value of  c-th channel $X_c$ at position$(i,j)$, $H_p$ is the global pooling function.
The shape of the feature map changes from $C\times H\times W$ to $C\times 1\times 1$. To get the weights of the different channels,  features pass through two convolution layers and sigmoid, ReLu  activation function latter.

$$CA_{c}=\sigma(Conv(\delta(Conv(g_{c})))) \eqno(4)$$

Where the $\sigma$ is the sigmoid function, $\delta $ is the ReLu function.

Finally, we element-wise multiply the input $F_c$ and the weights of the channel $CA_c$.

$$ F_{c}^{*}=CA_{c} \otimes F_{c} \eqno(5)$$

\subsubsection{Pixel Attention (PA)}
Considering that the haze distribution is uneven on the different image pixels, we propose a pixel attention (PA) module to make the network pay more attention to informative features, such as thick-hazed pixels and high-frequency image region. 

Similar to CA,  We directly feed the input $F^*$ (the output of the CA)  into two convolution  layers with ReLu and sigmoid activation function. The shape changes from $C\times H \times W$ to $1\times H\times W$. 

$$    PA=\sigma(Conv(\delta(Conv(F^{*})))) \eqno(6)$$

Finally we utilize element-wise multiplication for input $F^*$ and $PA$,  $\widetilde{F} $ is the output of the Future Attention (FA) module.
$$    \widetilde{F}=F^{*} \otimes PA \eqno(7)$$

\begin{figure}[t]
	\centering
	\subfigure[Pixel attention map]{
		\includegraphics[width=0.2\textwidth]{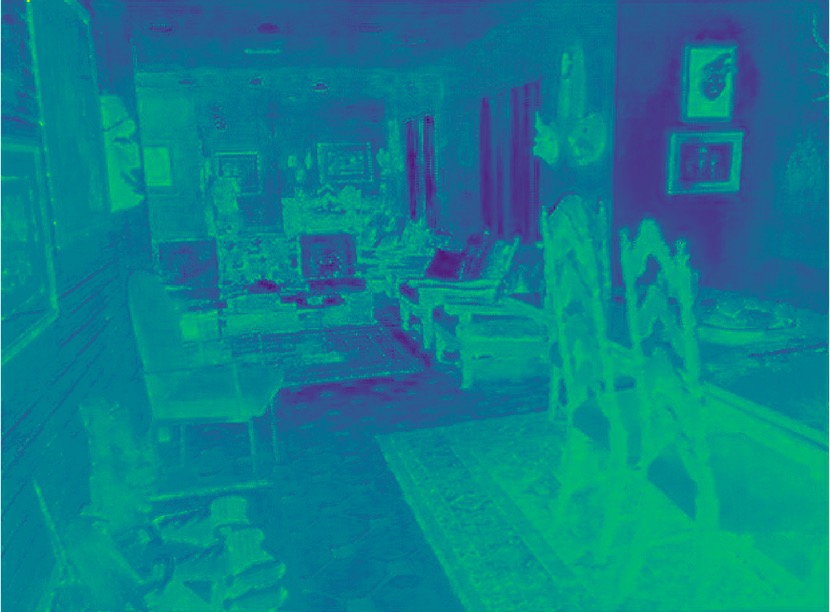} 
	}
	\subfigure[Input hazy image]{
		\includegraphics[width=0.2\textwidth]{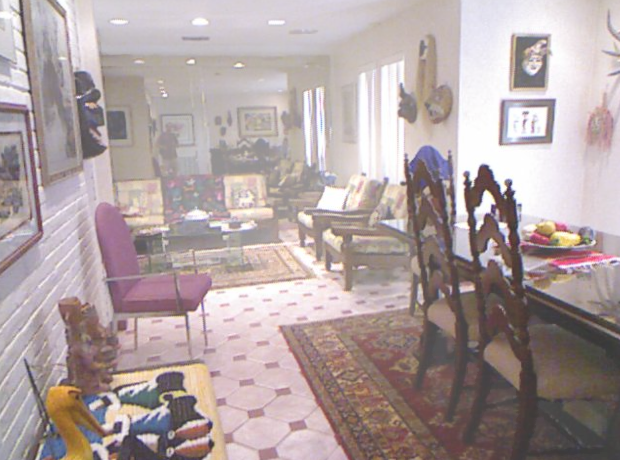}
	}
	\caption{PA attention map}

		\label{fig4}
				
\end{figure}

\begin{figure}[t]
	\centering
	\includegraphics[width=0.95\columnwidth]{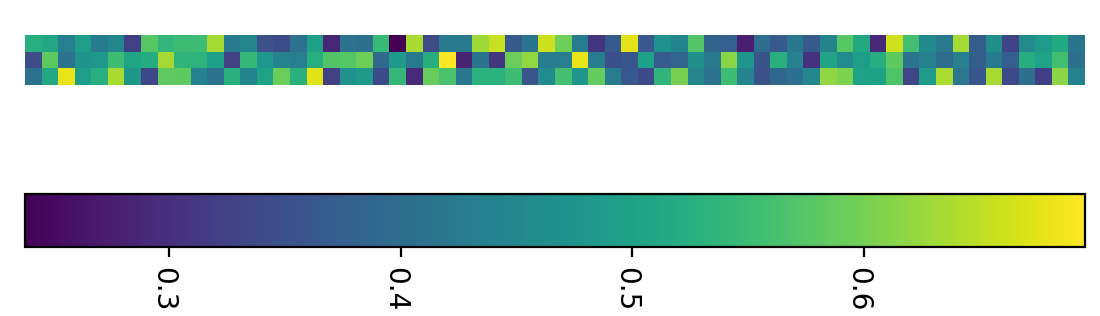} 
	\caption{Channel Attention weight map}
	\label{fig5}
\end{figure}

To visually illustrate the effectiveness of the feature attention (FA) mechanism, we print the channel-wise and pixel-wise feature weights map of the Group Structure output.
We can clearly see that different feature maps are adaptively learned with different weights.
Fig.\ref{fig4} shows that  thick hazy image pixel region and  the edges,  textures of objects  having a  larger weight. Pixel Attention (PA) mechanism makes the FFA-Net focus more attention on high frequency and thick-hazed pixels region.
Fig.\ref{fig5} shows a $3\times 64$ sized graph, and three rows correspond to the feature map weights of the  three Group Architectures output in the channel direction, illustration shows that different features  adaptively learn completely different weights.

\begin{figure}[t]
	\centering
	\includegraphics[width=0.95\columnwidth]{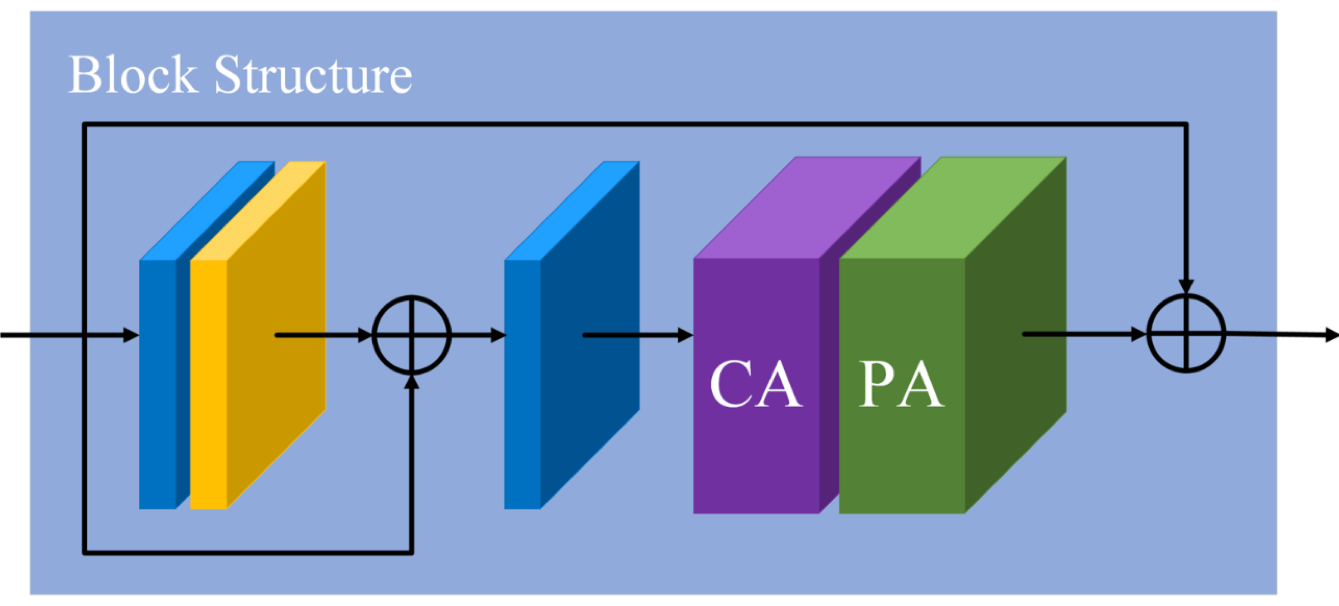} 
	\caption{ Basic Block Structure}
	\label{fig6}
\end{figure}

\subsection{Basic Block Structure}

As is shown in Fig.\ref{fig6}, a basic block structure consists of local residual learning and feature attention (FA) module, local residual learning allows the less important information such as thin haze or low-frequency region to be bypassed through multiple local residual connection, and main network focus on effective information. 

\begin{figure}[t]
	\centering
	\includegraphics[width=0.95\columnwidth,height=0.15\textheight]{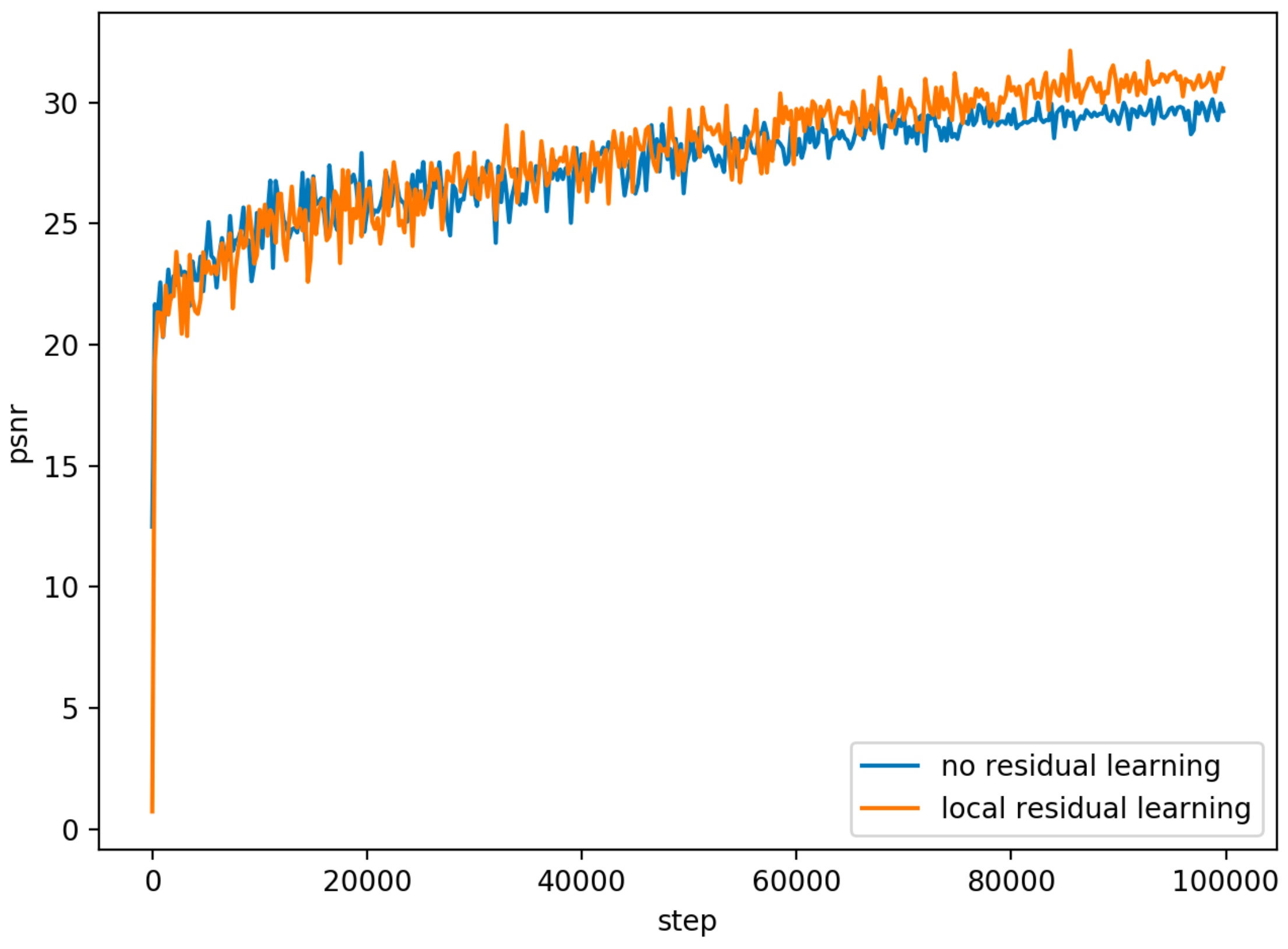} 
	\caption{The effect of local residual learning }
	\label{fig7}
\end{figure}

Experiment results show that its structure can further improve network performance and training stabilization, the effect of local residual learning can be seen in Fig.\ref{fig7}, specific details can be seen in the ablation study section

\subsection{Group Architecture and Global Residual Learning}
Our Group Architecture combines  B Basic Block structures with skip connections module. Continuous B blocks increase the depth and expressiveness of the FFA-Net. And skip connections make FFA-Net  get around training difficulty. 
At the end of the FFA-Net, we add a recovery part using a two-layer convolutional network implementation and a long shortcut global residual learning module. Finally, we  restore our desired haze-free image.
\subsection{Feature Fusion Attention }
As discussed above, firstly we concatenate all feature maps output by G Group Architectures in the channel direction. Furthermore, We fuse features by multiplying the adaptive learning weights which are obtained by Feature Attention (FA) mechanism. From this, we can retain the low-level information and pass it into deep layers, we let FFA-Net pay more attention to effective information such as thick haze region, high-frequency texture and color fidelity because of the weight mechanism.

\subsection{Loss Function}
Mean squared error (MSE) or L2 loss is the most widely used loss function for single image dehazing. However \cite{edsr}  pointed out that many image restoration tasks training with L1 loss achieved a better performance than L2 loss in terms of PSNR and SSIM metrics. Following the same strategy, we adopt the simple L1 loss by default. Although many dehazing algorithms also use the perceptual loss and GAN loss, we  choose to optimize the L1 loss.

$$ L(\Theta)=\frac{1}{N}\sum_{i=1}^{N}\| I_{gt}^{i}-FFA(I_{haze}^{i}) \| \eqno(8)$$
Where $\Theta$ denotes the parameters of FFA-Net, $	I_{gt}$ stands for ground truth, and $I_{haze}$ stands for input.

\subsection{Implementation Details}
In this section, we specify the implementation details of our proposed FFA-Net. The number of Group Structure  $G$ is 3. In each Group Structure, we set the Basic Block Structure number as $B=19$. Except for the Channel Attention whose kernel size is $1\times1$, we set all convolution layers’ filter size is $ 3\times 3$. All feature maps keep size fixed except for Channel Attention module. Every Group Structure outputs 64 filters.

\section{Experiments}

\subsection{Datasets and Metrics}

\cite{RESIDEbenchmarking}  proposed an image dehazing benchmark RESIDE, which contains synthetic hazy images  in both indoor and outdoor scenarios from depth dataset (NYU Depth V2\cite{nyu}) and
 stereo datasets(Middlebury Stereo datasets \cite{stereo}). 
 The Indoor Training Set of RESIDE contains 1399 clean image and 13990 hazy images generated by corresponding clean images. The global atmosphere light ranges from 0.8 to 1.0, the scatter parameters change from 0.04 to 0.2. To compare with previous state-of-the-art methods, we adopt PSNR and SSIM metrics and comprehensive comparisons testing in Synthetic Objective Testing Set (SOTS), which contains 500 indoor images and 500 outdoors ones. We also test the results on Realistic Hazy Images for subjective assessment.

\subsection{Training Settings}
We train the FFA-Net in RGB channels and augment the training dataset with randomly rotated by $90$,$180$,$270$ degrees and horizontal flip.
The $2$  hazy-image patches with the size $240\times 240$ are extracted as FFA-Net’s input. The whole network is trained for $5\times 10^5$, $1\times 10^6$ steps on indoor and outdoor images respectively. We use Adam optimizer, where $\beta 1$ and $\beta 2$ take the default values of 0.9 and 0.999, respectively. 

The initial learning rate is set to $1\times 10^{-4}$, 
we adopt the cosine annealing strategy \cite{cos} to adjust the learning rate from the initial value to 0 
by following the cosine function. Assume the total number of batches is $T$,$\eta$ is the initial
earning rate,  then at batch $t$, the learning rate $ \eta_{t}$ is computed as:
$$ \eta_{t}=\frac{1}{2}(1+\cos(\frac{t\pi}{T}))\eta \eqno(9)$$

  PytTorch \cite{pytorch}  was used to implement our models with a RTX 2080Ti GPU.
\begin{figure*}[!]
	\centering
	\subfigure[Indoor and outdoor results]{
		\includegraphics[width=0.9\textwidth,height=0.6\textheight]{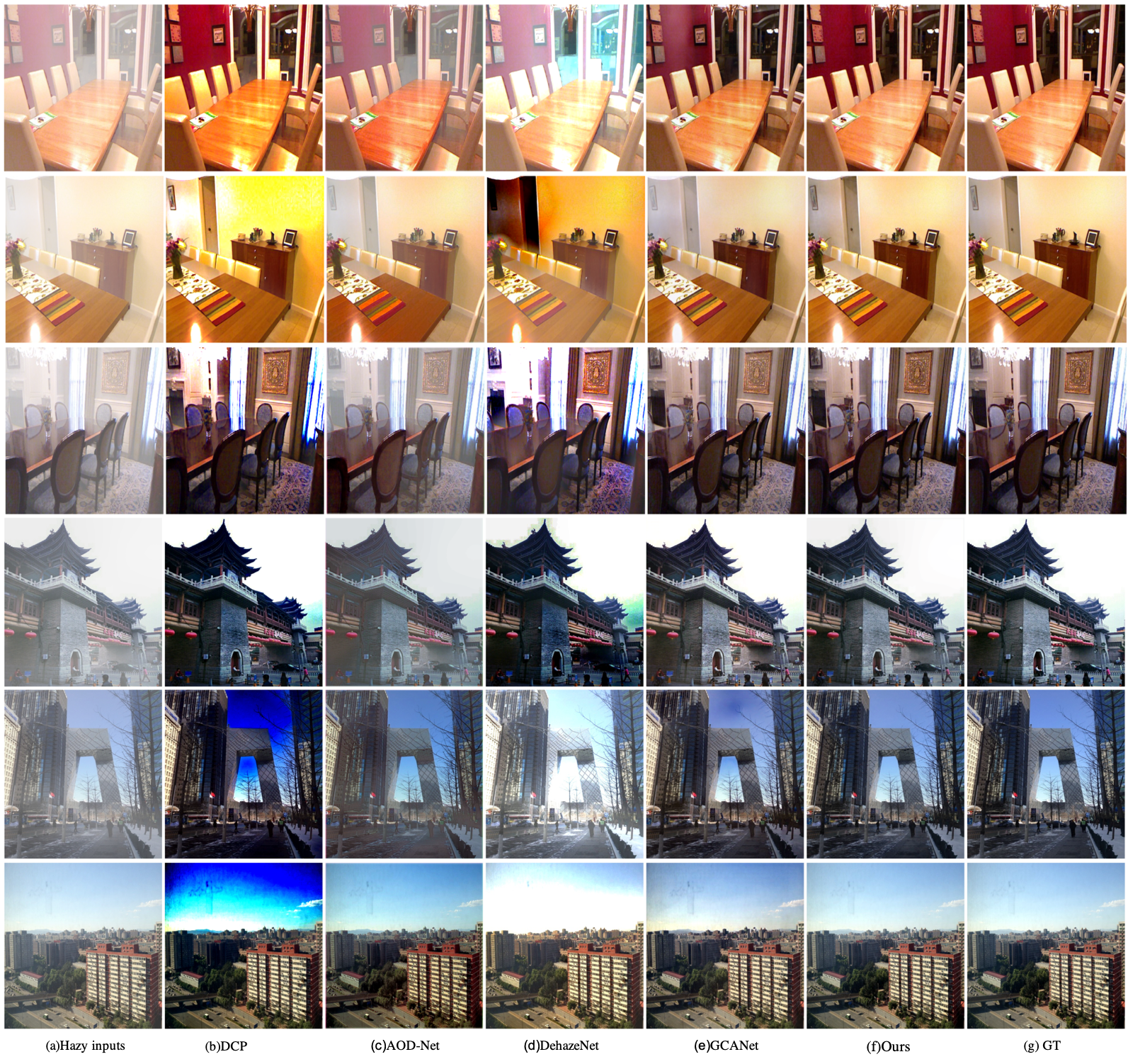} 
	}
	\subfigure[Real hazy image results]{
		\includegraphics[width=0.9\textwidth,height=0.2\textheight]{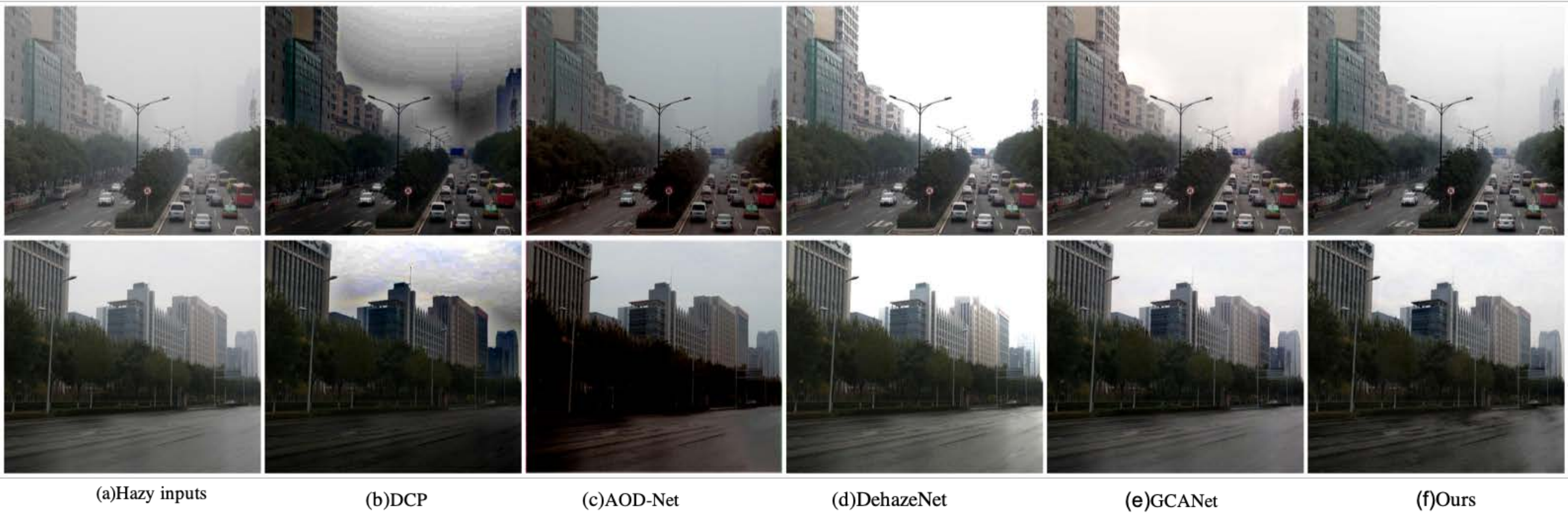}
	}

	\caption{Qualitative comparisons on SOTS and Realistic Hazy Images testset}

	\label{fig8}
	
\end{figure*}

\subsection{Results on RESIDE Dataset}

In this section, we will compare FFA-Net with previous state-of-the-art image dehazing algorithms both quantitatively and qualitatively.

We compared with  four different state-of-the-art dehazing algorithms which are the DCP, AOD-Net, DehazeNet, GCANet.The comparison results are shown in Table \ref{tab1} . 

 For convenience, the metrics are cited from \cite{RESIDEbenchmarking} and \cite{epdn}. It can be seen that our proposed 
 \begin{table}[h]
 	
 	\centering
 	
 	\begin{threeparttable}
 		
 		\begin{tabular}{ccccccc}
 			\toprule
 			{Methods}&
 			\multicolumn{2}{c}{ Indoor}&\multicolumn{2}{c}{ Outdoor}\cr
 			\cmidrule(lr){2-3} \cmidrule(lr){4-5}
 			&PSNR&SSIM&PSNR&SSIM\cr
 			\midrule
 			DCP&16.62&0.8179&19.13&0.8148\cr
 			AOD-Net&19.06&0.8504&20.29&0.8765\cr
 			DehazeNet&21.14&0.8472&22.46&0.8514\cr
 			GFN&22.30&0.8800&21.55&0.8444\cr
 			GCANet&30.23&0.9800&-&-\cr
 			\bf Ours&{\bf 36.39}&{\bf 0.9886}&{\bf 33.57}&{\bf 0.9840}\cr
 			\bottomrule
 		\end{tabular}
 		\caption{Quantitative comparisons on SOTS for different methods.}
 		\label{tab1}
 	\end{threeparttable}
 \end{table}
FFA-Net outperforms all four different state-of-the-art methods by a very large margin in terms of PSNR and SSIM.
Moreover, we give the comparison of the visual effect in Fig.\ref{fig8} for qualitative comparisons.

From the indoor and outdoor results,  three upper rows are indoor results, and outdoor results  are shown in the bottom three rows. we can observe that DCP suffers from severe color distortion  because of  their underlying prior assumptions, consequently, it loses the details in the depth of image. AOD-Net  can not remove the haze completely and  tends 
to output low-brightness images. In contrast, Dehazenet recovers images with excessive brightness relative to ground truth.  The processing power of GCANet at high-frequency detail  information performance such as textures, edges and the blue sky in row 5 is always unsatisfactory.

For real hazy image results, our network can magically discover the towers that are looming in the depths of the image in row 1. More importantly, the results of our network  are almost entirely in line with real  scene information, such as the wet  road with texture and raindrops in row 2.
However, it is found that there are nonexistent spots on the building surface of the GCANet result in row 2. The images recovered from other networks are not satisfactory.
Our network is clearly superior in the realistic performance of image details and color fidelity.


\section{Ablation Analysis}
To further demonstrate the superiority of FFA-Net architecture, we conduct an ablation study by considering the different modules of our proposed FFA-Net. We mainly concern  these factors: 1)  The FA  (Feature Attention) module. 2)  The combination of local residual learning  (LRL)  and FA.  3)  The Feature Fusion Attention (FFA) structure.
We crop the image to $48\times 48$ as input with training of $3\times 10^5 $ steps, other configuration is the same as our implementation details. 
The results are shown in Table \ref{tab2}.

\begin{table}[h]
	
	\centering
	
	\begin{threeparttable}
		
		\begin{tabular}{cccccc}
			\toprule
			FA&\checkmark &\checkmark&\checkmark \cr
			LRL& &\checkmark&\checkmark \cr
			FFA& & &\checkmark& \cr
			\midrule
			PSNR &30.28db &31.16db &32.44db \cr
			\bottomrule
		\end{tabular}
		\caption{Comparisons on SOTS indoor testset for different configurations.}
		\label{tab2}
	\end{threeparttable}
\end{table}

And If we fully use the implementation details in our paper, then PSNR will achieve 35.77db.
The results show that every factor we consider plays an important role in   the network performance,  especially the FFA structure. We can also clearly see that even if we only use the FA structure, our network can be very competitive compared with previous state-of-the-art methods. LRL makes network training stable while improving the network performance.  The combination of the FA mechanism and feature fusion (FFA) has brought our results to a very high level.

\section{Conclusion}
In this paper, we propose an end-to-end Feature Fusion Attention Network and demonstrated its strong power in single image dehazing. Although Our FFA-Net structure is simple, it is better than the previous state-of-the-art methods with a very large margin. Our network has a powerful advantage  in the restoration of image detail and color fidelity, it is expected to solve other low-level vision tasks such as deraining,  super-resolution, denoising. FFA and other effective modules in our FFA-Net play an important role in the image restoration algorithms.

 \textbf{Acknowledgements.} This work is partially supported by the National Key Research and Development Program of China under contract No. 2016YFB0402001.

\bibliographystyle{aaai}
\bibliography{ref}

\end{document}